\documentclass{article} 
\usepackage{graphicx}
\usepackage{pgfplots}
\usepackage{amsmath}
\usepackage{colortbl}
\usepackage[left=4cm,right=4cm]{geometry}

\usetikzlibrary{patterns}

\title{Mixtures of conditional Gaussian scale mixtures \\
applied to multiscale image representations}

\author{
	Lucas Theis \quad Reshad Hosseini \quad Matthias Bethge \\
	Werner Reichardt Center for Integrative Neuroscience \\
	Max Planck Institute for Biological Cybernetics \\
	Spemannstraße 41, 72076 Tübingen, Germany \\
	\texttt{\{lucas,hosseini,mbethge\}@tuebingen.mpg.de} \\
}

\begin{document}
	\maketitle

	\begin{abstract}
		We present a probabilistic model for natural images which is based on
		Gaussian scale mixtures and a simple multiscale representation. In contrast to the
		dominant approach to modeling whole images focusing on Markov random
		fields, we formulate our model in terms of a directed graphical model.
		We show that it is able to generate images with interesting higher-order
		correlations when trained on natural images or samples from an occlusion
		based model. More importantly, the directed model enables us to perform
		a principled evaluation. While it is easy to generate visually
		appealing images, we demonstrate that our model also yields
		the best performance reported to date when evaluated with respect to the
		cross-entropy rate, a measure tightly linked to the average
		log-likelihood.
	\end{abstract}

	\section{Introduction}
		Probabilistic models of natural images are used in many fields related
		to vision. In computational neuroscience, they are used as a means to
		understand the structure of the input to which biological vision systems
		have adapted and as a basis for normative theories of how those inputs
		are optimally processed~\cite{Gallant:6505266}. In computer science,
		they are used as priors in applications such as image
		denoising~\cite{GuerreroColon:2008p7482},
		compression~\cite{Hosseini:2007}, or reconstruction
		\cite{Domke:2008p7554} and to learn image representations that can be
		used in object recognition tasks~\cite{Ranzato:2010p7468}. The more
		abstract goal common to these efforts is to capture the statistics of
		natural images.

		The dominant approach to modeling whole images has been to use
		undirected graphical models (or \textit{Markov random fields}). This is despite
		the fact that directed models possess many advantages over undirected
		models \cite{Domke:2008p7554,Hosseini:2010p7310}. In particular,
		sampling as well as exact maximum likelihood learning can often be
		performed efficiently in directed models while presenting a major
		challenge with most undirected models. Another problem faced by
		undirected models is the question of how to evaluate them. Ideally, we
		would like to quantify the amount of second- and higher-order
		correlations captured by a model. For stochastic processes, this can be
		done by calculating the cross-entropy rate between the learned
		distribution and the true distribution, which represents the analogue to
		the negative log-likelihood for patch-based models. However, the
		cross-entropy rate is typically difficult to estimate in undirected
		models so that these models are often evaluated only with respect to surrogate
		measures such as performance in supervised tasks, simple statistics
		computed from approximate model samples or simply the samples' visual
		appearance. These measures, however, are less
		objective and hence need to be used with great caution. A large lookup
		table storing examples from the training set, for example, will
		reproduce samples which are indistinguishable from true samples for the
		human eye and yield near to perfect performance when evaluated based on
		simple statistics. Yet this model is heavily overfit to a few examples
		of natural images. Effectively, it assigns zero probability to images
		that have not been stored in the lookup table and would thus perform
		miserably if evaluated based on the cross-entropy rate. In fact, the
		cross-entropy rate does not leave any room for prestidigitation and therefore
		provides a crucial basis for the comparison of natural image models.

		Following the directed approach, we will demonstrate here that a
		directed model applied to multi-scale representations of natural images
		is able to learn and reproduce interesting higher-order correlations. We
		use multiscale representations to separate the coarser components of an
		image from its details, thereby facilitating the modeling of both very
		global and very local image structure. The particular choice of our
		representation makes it possible to still evaluate the cross-entropy rate
		and to further investigate the scale-invariance of natural images.

	\section{A directed model for natural images}
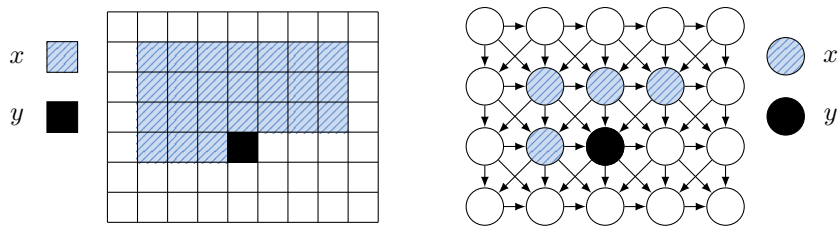
\begin{figure}[t]
	\centering
	\begin{tikzpicture}[
			scale=0.4,
			nbbg/.style={
				fill=blue!70!green!20,
				inner sep=5.8
			},
			nbpt/.style={
				pattern color=blue!70!green!60,
				pattern=north east lines,
				inner sep=5.8
			},
			unit/.style={
				circle, draw, inner sep=5
			},
			unbg/.style={
				fill=blue!70!green!20,
				circle, inner sep=5
			},
			unpt/.style={
				pattern=north east lines,
				pattern color=blue!70!green!60,
				circle, inner sep=5
			}
		]
		\begin{scope}
			\foreach \x in {0,...,6}
				\foreach \y in {4,...,6} {
					\node [nbbg] at (\x + 0.5, \y + 0.5) {};
					\node [nbpt] at (\x + 0.48, \y + 0.475) {};
				}

			\foreach \x in {0,...,2} {
				\node [nbbg] at (\x + 0.5, 3.5) {};
				\node [nbpt] at (\x + 0.48, 3.48) {};
			}

			\node [fill=black, inner sep=5.8] at (3.5, 3.5) {};

			\foreach \x in {-1,...,7} {
				\foreach \y in {1,...,7} {
					\draw (\x, \y) -- (\x + 1, \y);
					\draw (\x, \y) -- (\x, \y + 1);
				}
				\draw (\x, 8) -- (\x + 1, 8);
			}
			\foreach \y in {1,...,7}
				\draw (8, \y) -- (8, \y + 1);

			\node [fill=black, draw, inner sep=5.8] at (-2.5, 4.5) {};
			\node [nbbg] at (-2.5, 6.5) {};
			\node [nbpt, draw] at (-2.5, 6.5) {};
			\node at (-4, 4.5) {$y$};
			\node at (-4, 6.5) {$x$};
		\end{scope}

		\begin{scope}[xshift=300,yshift=-42]
			\node [unbg] (unit37) at (3, 7) {};
			\node [unpt] (unit37) at (3, 7) {};
			\node [unbg] (unit57) at (5, 7) {};
			\node [unpt] (unit57) at (5, 7) {};
			\node [unbg] (unit77) at (7, 7) {};
			\node [unpt] (unit77) at (7, 7) {};
			\node [unbg] (unit35) at (3, 5) {};
			\node [unpt] (unit35) at (3, 5) {};
			\node [unit, fill=black] (unit55) at (5, 5) {};

			\foreach \x in {1,3,5,7,9}
				\foreach \y in {3,5,7,9}
					\node [unit] (unit\x\y) at (\x, \y) {};

			\foreach \x in {1,3,5,7} {
				\pgfmathparse{\x + 2};
				\pgfmathtruncatemacro\p{\pgfmathresult};
				\foreach \y in {5,7,9} {
					\pgfmathparse{\y - 2};
					\pgfmathtruncatemacro\q{\pgfmathresult};
					\draw [-latex] (unit\x\y) to (unit\p\y);
					\draw [-latex] (unit\x\y) to (unit\x\q);
					\draw [-latex] (unit\x\y) to (unit\p\q);
					\draw [-latex] (unit\p\y) to (unit\x\q);
				}
				\draw [-latex] (unit\x3) to (unit\p3);
			}

			\foreach \y in {5,7,9} {
				\pgfmathparse{\y - 2};
				\pgfmathtruncatemacro\q{\pgfmathresult};
				\draw [-latex] (unit9\y) to (unit9\q);
			}

			\node [unbg] (unit37) at (11, 8) {};
			\node [unpt, draw] (unit37) at (11, 8) {};
			\node [unit, fill=black] (unit55) at (11, 6) {};
			\node at (12.5, 6) {$y$};
			\node at (12.5, 8) {$x$};
		\end{scope}
	\end{tikzpicture}
	\caption{
		\textit{Left:} A conditional model with a twenty-four pixel causal
		neighborhood. Sampling is performed by shifting the causal neighborhood
		from left to right and from top to bottom. \textit{Right:} Graphical model
		representation with only four pixels in the causal neighborhood.
		The parents of a pixel are constrained to pixels which are above of it
		or in the same row and left of it.
	}
	\label{fig:directedmodel}
\end{figure}
		One way to model the statistics of arbitrarily large images is to use a
		directed model in which the parents of a node are constrained to pixels
		which are left or above of it (Figure~\ref{fig:directedmodel}).
		A set of parents fulfilling this constraint is also called a
		\textit{causal neighborhood} \cite{Hosseini:2010p7310} and the resulting model a
		\textit{causal random field}. Note that a
		pixel will still depend on neighbors in all directions, that is, the
		causal neighborhood assumption puts only mild constraints on the size or
		shape of a pixel's Markov blanket. An advantage of the directed
		model is that it allows us to easily decompose the
		distribution defined over images or, more generally, a two-dimensional
		stochastic process $X$ indexed by $x, y$, into a product of conditional distributions:
		\begin{align}
			P(X) = \prod_{x, y} P(X_{x, y} \mid \mathrm{Pa}_{x, y}),
			\label{eq:directed_model}
		\end{align}
		where $\mathrm{Pa}_{x, y}$ refers to the causal neighborhood of pixel $X_{x, y}$.
		Consequently, performing maximum likelihood learning by maximizing the
		log-likelihood of the model can be done by optimizing a set of
		conditional probability distributions. To sample an image from the
		model, the causal neighborhood is shifted from top to bottom and from
		left to right, filling an image row by row.
		This procedure requires that the top rows and left columns of the image are
		initialized to provide input to the conditional distributions. Generally,
		these cannot be filled with pixels drawn from the distribution of the
		model. As a consequence, only after the procedure has generated a few
		rows and converged to the model's distribution will it
		generate the desired samples.

	\subsection{Mixture of conditional Gaussian scale mixtures}
		To complete the model, the conditional distribution of each pixel given
		its causal neighborhood has to be specified. We will assume stationarity
		(or shift-invariance), so that this task reduces to the specification of
		a single conditional distribution. A family of distributions which has
		repeatedly been shown to contain suitable building blocks for modeling
		the statistics of natural images is given by \textit{Gaussian scale
		mixtures} (GSMs)
		\cite{Wainwright:2000p7480,Weiss:2007p7306},
		\begin{align}
			p(x) = \int \varphi(z) \mathcal{N}(x; \mu, z C) dz,
		\end{align}
		where $\mathcal{N}(x; \mu, z C)$ is a multivariate Gaussian density with
		mean $\mu$ and covariance $z C$ and $\varphi(z)$ is some univariate
		density over scales $z$. Mixture models and Markov
		random fields based on GSMs have been successfully applied to denoising
		tasks \cite{GuerreroColon:2008p7482,Lyu:2007p7424}. When used in the
		directed setting also employed here, GSMs have been shown to yield highly
		improved estimates of the multi-information rate of natural images \cite{Hosseini:2010p7310}.

		Here we use the conditional distribution of a mixture of GSMs to model
		the distribution of a pixel given its causal neighborhood. We restrict ourselves to
		mixtures of finite GSMs, that is, GSMs with a
		finite number of scales, and to mixtures in which each component and
		scale has equal a priori weight. Additionally, we assume that each
		GSM has mean zero. If variables $x$ and $y$ are modeled jointly with
		a mixture of GSMs, the conditional distribution of $y$ given $x$ can be written as
		\begin{align}
			p(y \mid x) &= \sum_{c, s} \underbrace{p(c, s \mid x)}_\text{gate} \underbrace{p(y \mid x, c, s)}_\text{expert},
		\end{align}
		where $c, s$ run over mixture components and scales, respectively. From this
		formulation it is clear that the conditional distribution falls into the
		\textit{mixtures of experts} framework \cite{Jacobs:1991p7614}. In this framework, the predictions of
		multiple predictors---the \textit{experts}---are mixed according to weights which are
		computed locally by so called \textit{gates}. For the mixture of GSMs
		with the constraints above we have
		\begin{align}
			p(c, s \mid x) &\propto |\lambda_{cs} K_c|^\frac{1}{2} \exp(-\frac{1}{2} x^\top \lambda_{cs} K_c x), \\
			p(y \mid x, c, s) &\propto \exp(-\frac{1}{2} (y - A_c x)^\top \lambda_{cs} M_c (y - A_c x)),
		\end{align}
		where $M_c$ and $K_c$ are positive definite matrices and $\lambda_{cs}$ are
		positive. The gates pick an expert based on the covariance
		structure and scale of the input variables $x$. Each expert is just a
		Gaussian with a linearly predicted mean. The conditional distribution
		can equivalently be described as a mixture of conditional Gaussian scale
		mixtures (MCGSM), because conditioned on $c$ the conditional
		distribution is again a GSM.

	\subsection{A simple multiscale representation}
\begin{figure}[t]
	\centering
	\begin{tikzpicture}[
			nbbg/.style={
				fill=blue!70!green!20,
				color=blue!70!green!20,
				line/.style={loosely dotted},
				line width=0pt,
			},
			nbpt/.style={
				pattern color=blue!70!green!60,
				pattern=north east lines,
				line/.style={loosely dotted},
				line width=0pt,
			}
		]
		\draw[step=.2,very thin,gray] (-1.6,-1.6) grid (1.6,1.6);
		\draw[step=.4] (-1.6,-1.6) grid (1.6,1.6);

		\draw [-latex] (1.7, 0) -- (2.7, 0);

		\begin{scope}[xshift=108,yshift=-3.5]
			\draw[xshift=14,   yshift=12,   fill=blue!70!green!10]   (-0.8, -0.8) rectangle (0.8, 0.8);
			\draw[xshift=14,   yshift=12, nbbg]         (0.2, 0.0) rectangle (0.8, 0.6);
			\draw[xshift=14,   yshift=12, nbpt]         (0.2, 0.0) rectangle (0.8, 0.6);
			\draw[xshift=14,   yshift=12,   step=0.2, very thin]             (-0.8, -0.8) grid      (0.8, 0.8);

			\draw[xshift=3.5,  yshift=3.5,  fill=white]                      (-0.8, -0.8) rectangle (0.8, 0.8);
			\draw[xshift=3.5,  yshift=3.5, nbbg]         (0.2, 0.4) rectangle (0.8, 0.6);
			\draw[xshift=3.5,  yshift=3.5, nbpt]         (0.2, 0.4) rectangle (0.8, 0.6);
			\draw[xshift=3.5,  yshift=3.5,  step=0.2, very thin]             (-0.8, -0.8) grid      (0.8, 0.8);

			\draw[xshift=0,    yshift=0,    fill=white]                      (-0.8, -0.8) rectangle (0.8, 0.8);
			\draw[xshift=0,    yshift=0, nbbg]         (0.2, 0.4) rectangle (0.8, 0.6);
			\draw[xshift=0,    yshift=0, nbpt]         (0.2, 0.4) rectangle (0.8, 0.6);
			\draw[xshift=0,    yshift=0,    step=0.2, very thin]             (-0.8, -0.8) grid      (0.8, 0.8);

			\draw[xshift=-3.5, yshift=-3.5, fill=white]                      (-0.8, -0.8) rectangle (0.8, 0.8);
			\draw[xshift=-3.5, yshift=-3.5, nbbg]         (0.2, 0.4) rectangle (0.8, 0.6);
			\draw[xshift=-3.5, yshift=-3.5, nbpt]         (0.2, 0.4) rectangle (0.8, 0.6);
			\draw[xshift=-3.5, yshift=-3.5, nbbg]         (0.2, 0.2) rectangle (0.4, 0.4);
			\draw[xshift=-3.5, yshift=-3.5, nbpt]         (0.2, 0.2) rectangle (0.4, 0.4);
			\draw[xshift=-3.5, yshift=-3.5, step=0.2, very thin]             (-0.8, -0.8) grid      (0.8, 0.8);
			\draw[xshift=-3.5, yshift=-3.5, very thin, fill=black]        (0.4, 0.2) rectangle (0.6, 0.4);
		\end{scope}

		\node at (2.2, 0.35) {$\phi$};
		\node at (0, 2) {pixel representation};
		\node at (0, -2) {$X_0$};
		\node at (4, 2.2) {superpixel};
		\node at (3.7, -1.5) {$Y_1$};
		\node at (5.4, -0.5) {$X_1$};
		\node at (4, 1.8) {representation};
		\node at (8, 2) {Haar wavelet basis};

		\begin{scope}[xshift=5]
			\begin{scope}[xshift=-15,yshift=15]
				\node [draw, inner sep=5.8, fill=black] at (8.2025,  0.2025) {};
				\node [draw, inner sep=5.8, fill=black] at (7.7975,  0.2025) {};
				\node [draw, inner sep=5.8, fill=white] at (8.2025, -0.2025) {};
				\node [draw, inner sep=5.8, fill=white] at (7.7975, -0.2025) {};
			\end{scope}

			\begin{scope}[xshift=15,yshift=15]
				\node [draw, inner sep=5.8, fill=black] at (8.2025,  0.2025) {};
				\node [draw, inner sep=5.8, fill=white] at (7.7975,  0.2025) {};
				\node [draw, inner sep=5.8, fill=black] at (8.2025, -0.2025) {};
				\node [draw, inner sep=5.8, fill=white] at (7.7975, -0.2025) {};
			\end{scope}

			\begin{scope}[xshift=-15,yshift=-15]
				\node [draw, inner sep=5.8, fill=white] at (8.2025,  0.2025) {};
				\node [draw, inner sep=5.8, fill=white] at (7.7975,  0.2025) {};
				\node [draw, inner sep=5.8, fill=white] at (8.2025, -0.2025) {};
				\node [draw, inner sep=5.8, fill=white] at (7.7975, -0.2025) {};

				\node at (6.5, 0.8) {\scriptsize low resolution};
				\node at (6.5, 0.5) {\scriptsize channel};
				
				\draw [latex-] (7.5, 0.0) parabola (6.8, 0.3);
				\draw [latex-] (5.5, 1.5) parabola (6.25, 1.0);
			\end{scope}

			\begin{scope}[xshift=15,yshift=-15]
				\node [draw, inner sep=5.8, fill=black] at (8.2025,  0.2025) {};
				\node [draw, inner sep=5.8, fill=white] at (7.7975,  0.2025) {};
				\node [draw, inner sep=5.8, fill=white] at (8.2025, -0.2025) {};
				\node [draw, inner sep=5.8, fill=black] at (7.7975, -0.2025) {};
			\end{scope}
		\end{scope}
	\end{tikzpicture}
	\caption{
		Starting with a regular grey-scale image, the pixels are grouped into two by two pixels.
		Each group is then transformed using the Haar wavelet basis on the right.
		The resulting basis coefficients can be interpreted as channels of an image of
		which one channel is essentially the original image at a coarser scale (lower
		resolution). Just as in the original representation, we can define a
		directed model and causal neighborhoods for the superpixel
		representation. If the low-resolution image is given,
		the prediction of a pixel can be based on information from anywhere in the
		low-resolution image (not just a causal neighborhood) without losing
		the ability to efficiently sample or optimize the parameters of the model.
	}
	\label{fig:multiscale}
\end{figure}
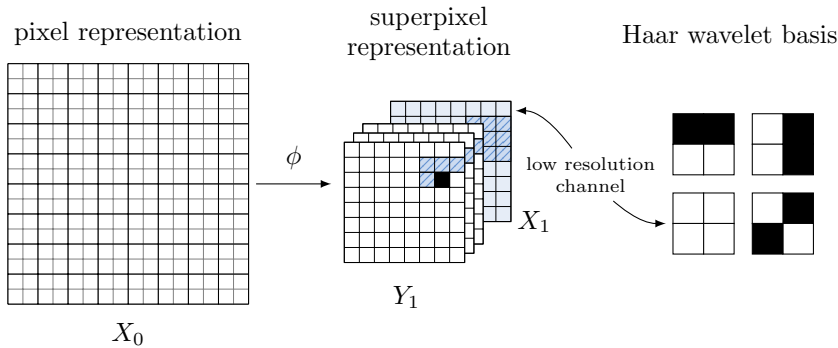

		To facilitate the modeling of global as well as local structure, we
		introduce a multiscale representation which allows us to model an image
		at multiple resolutions. We start by
		transforming an image into a representation consisting of a
		low-resolution version of the image and a separate representation for
		its details. This is achieved by grouping the image
		into patches of two by two pixels and transforming each patch using a
		Haar wavelet basis. One component of the Haar wavelet basis
		essentially performs a block-average on the image, the result of which
		forms the low-resolution part of the representation.
		The three remaining coefficients encode the more detailed structure (Figure~\ref{fig:multiscale}).

		The transformed image can be viewed as an image with multiple channels
		where each set of basis coefficients corresponds to one channel. We will
		call the multi-channel pixels in the new representation \textit{superpixels}. Just
		as we would model RGB images, we can model images in this new
		representation with the MCGSM by predicting all channels of a superpixel
		at once.

		If we assume further that the low-resolution channel of the
		image is given, we can base our predictions on an arbitrary set of
		pixels from the low-resolution image without losing the ability to
		efficiently perform maximum likelihood learning or to sample from the
		model. This can be seen as follows.
		Just like the transformation of four pixels using the Haar wavelet basis,
		the transformation between the two representations, $\phi$, is a linear
		transformation with a Jacobian determinant of 1. If $\phi(X^0) =
		(Y^1, X^1)$ is the superpixel representation of an image $X^0$, consisting of a low-resolution part
		$X^1$ and a high-resolution part $Y^1$, we have
		\vspace{3pt}
		\begin{align}
			P(X^0)
			= P(\phi(X^0)) \underbrace{|\det D \phi(X^0)|^{-1}}_{= 1}
			= P(Y^1, X^1)
			= P(Y^1 \mid X^1) P(X^1).
		\end{align}
		Each factor on the right-hand side again factors into a product of the
		form of Equation \ref{eq:directed_model}.
		Sampling an image is achieved by first sampling a low-resolution image
		$X^1$ and then conditionally sampling~$Y^1$. Every variable that has already
		been sampled can be used to sample and predict the remaining variables.
		Maximizing the log-likelihood amounts to maximizing the logarithm of the two factors on the right-hand side.

		By recursively applying the transformation to the low-resolution image of
		the representation, we end up with a pyramid of images where each level contains
		the high-resolution information that completes the image represented by
		the levels above of it. At the top of the pyramid, that is, the lowest resolution, we will again model
		the image with an MCGSM, so that after applying the transformation $M$
		times we are using $M + 1$ MCGSMs.


	\subsection{Model evaluation}
		A principled way to evaluate a model approximating a stochastic process
		$X$ is to use the model to estimate the true distribution's
		\textit{multi-information rate} (MIR),
		\begin{align}
			I_\infty[X] = \lim_{N \rightarrow \infty} \frac{1}{N} \sum_{n = 1}^N H[X_n] - H[X_1, ..., X_N],
		\end{align}
		where $H$ denotes the (differential) entropy. A related measure is the \textit{entropy rate},
		\begin{align}
			H_\infty[X] = \lim_{N \rightarrow \infty} \frac{1}{N} H[X_1, ..., X_N].
		\end{align}
		For a strictly stationary Markov process, one can show that these
		quantities reduce to \cite{Hosseini:2010p7310,Cover:2006p7308}
		\begin{align}
			H_\infty[X] &= H[X_N \mid \mathrm{Pa}_N],
			\label{eq:entropy_rate} \\
			I_\infty[X] &= H[X_1] - H[X_N \mid \mathrm{Pa}_N],
			\label{eq:information_rate}
		\end{align}
		for some $N$. By replacing the entropy rate with the
		\textit{cross-entropy rate}---a limit of cross-entropies instead of
		entropies---we obtain a lower bound on the true MIR. We will call this
		lower bound the \textit{cross-MIR} in the following. If the
		assumption of stationarity or the Markov assumption is not met by the
		true distribution, the cross-MIR will still be a lower bound but
		will become more loose \cite{Hosseini:2010p7310}. The difference between
		the true MIR and the cross-MIR is the Kullback-Leibler divergence
		between the true distribution and the model distribution. Therefore,
		the better the approximation of the model to the true distribution, the
		larger the cross-MIR.

		Maximizing the cross-MIR by minimizing the cross-entropy rate is the
		same as maximizing the average log-likelihood. The MIR quantifies the amount of second- and
		higher-order correlations of a stochastic process.
		Similar to the likelihood, the cross-MIR can be said to capture the
		amount of correlations learned by the model. In addition, it has the
		advantage of being easier to interpret than the likelihood or the
		cross-entropy rate, as it is always non-negative and invariant under
		multiplication of the stochastic process with a constant factor. An
		independent white noise process has a MIR of zero.
		In the stationary case, evaluating the cross-MIR
		amounts to calculating one marginal entropy
		and one conditional cross-entropy (see Equation
		\ref{eq:information_rate}).

		Since the superpixel representation is just a linear
		transformation of the original image, we can evaluate the entropy rate
		also for the multiscale model. Using the fact that the transformation has a Jacobian
		determinant of~1, the following relationship holds for the entropy
		and cross-entropy rates:
		\begin{align}
			H_\infty[X^0] = \frac{1}{4} H_\infty[Y^1 \mid X^1] + \frac{1}{4}
			H_\infty[X^1].
			\label{eq:entropy_rate_splitting}
		\end{align}
		The factor $\frac{1}{4}$ is due to the superpixel representation
		having four channels. This result readily generalizes to the multiscale
		representation by recursively applying it to the right term on the
		right-hand side of Equation \ref{eq:entropy_rate_splitting}. This
		means that in order to estimate the cross-entropy rate of our model, we
		only need to compute the cross-entropy rates at the
		different scales and form a weighted average.

	\section{Experiments}
\begin{figure}[t]
	\centering
	\includegraphics[width=\textwidth]{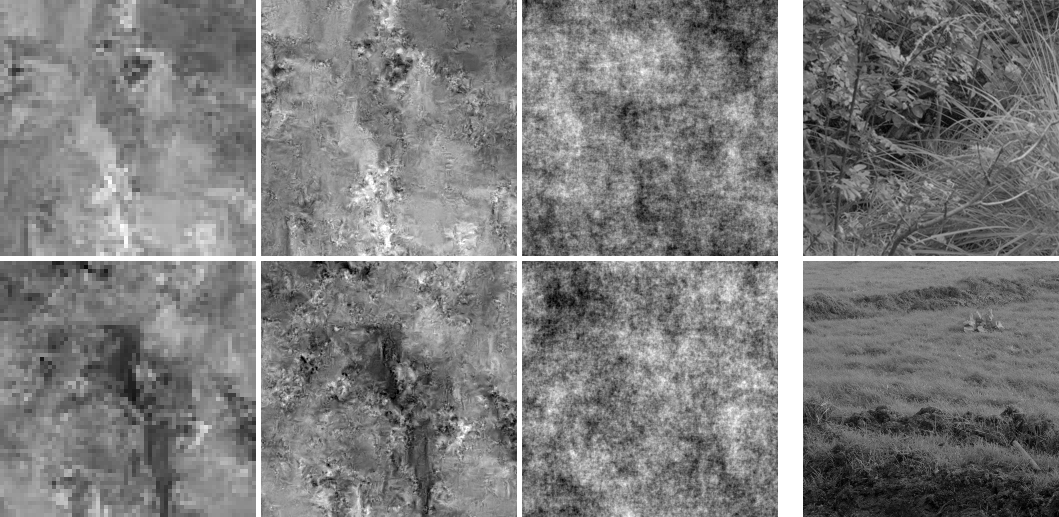}
	\caption{
		The first column shows a sample from the MCGSM at the largest scale.
		The second column shows a sample from our model at
		a smaller scale, conditionally sampled with respect to the sample on the
		left. The third column shows the same samples with all higher-order
		correlations destroyed but the autorocorrelation function left intact. The right
		most column shows examples of images from the training set, that is, the van
		Hateren dataset \cite{vanHateren:1998p7367}.
	}
	\label{fig:vanhateren_samples}
\end{figure}

		We extracted training data at four different scales from log-transformed
		images taken from the van Hateren dataset \cite{vanHateren:1998p7367}.
		In all experiments, we used $200000$ training examples of inputs and outputs.
		To model the coarsest scale, we used an MCGSM with a
		causal neighborhood corresponding to the upper half of a $7 \times
		7$ neighborhood surrounding the predicted pixel (as in Figure
		\ref{fig:directedmodel}). For the finer scales, we trained three MCGSMs
		with $3 \times 3$ superpixel neighborhoods (as in Figure
		\ref{fig:multiscale}). All models consisted of 8 components with 4 scales each.
		The parameters were trained using the BFGS quasi-Newton
		method \cite{JorgeNocedal:1999p7547}. For faster convergence, we initialized the conditional models
		with parameters from mixtures of GSMs trained on the joint distribution
		of inputs and outputs using expectation maximization. Initializing the
		models in this way also led to more stable as well as slightly
		better results. At each scale, we initialized the boundaries of the
		image with small random white noise and sampled an image large enough for the
		sampling procedure to reach convergence to the model's stationary
		distribution. We then extracted the center part of the image and used it
		as the input to the model at the next scale.

		Samples from the model are shown in Figure \ref{fig:vanhateren_samples}.
		We find that the model is able to generate images with some interesting
		properties which cannot be found in other models of natural images.
		Perhaps the most striking property of the sampled images is the
		heterogeneity expressed in the combination of flat image regions with
		regions of high variance as it can also be observed in true natural images.

		By destroying the higher-order correlations but keeping the second-order
		correlations, we get the familiar pink noise images (Figure
		\ref{fig:vanhateren_samples}). This shows that the model faithfully
		reproduces the autocorrelation function of natural images, and that the
		characteristic features of the sampled images are due to higher-order
		correlations learned by the model. The higher-order correlations were
		removed by replacing the phase spectrum of the
		image with a random phase spectrum obtained from a white noise image but
		keeping the sample's amplitude spectrum. For stationary processes, the
		amplitude spectrum defines the autocorrelation function of an image and
		vice versa.

\begin{figure}[t]
	\centering
	\includegraphics[width=\textwidth]{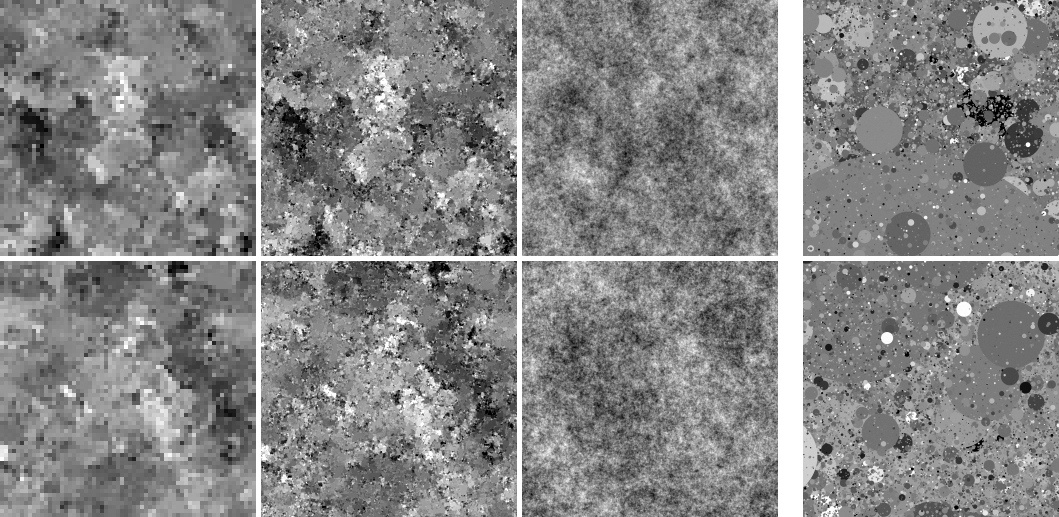}
	\caption{
		The model was trained on samples from an occlusion-based model
		\cite{Lee:1999p7586}. Example images from the training set are given on
		the right.  As above, the first two columns show samples from our model
		at two different scales. The third column shows the same samples with
		all higher-order correlations destroyed, revealing second-order
		statistics which are very similar to the ones learned from natural
		images.
	}
	\label{fig:leaves_samples}
\end{figure}

		As another test, we generated 1000 images of size $256 \times 256$
		pixels from an occlusion model ("\textit{dead leaves}") using the
		procedure described by Lee and Mumford in \cite{Lee:1999p7586}.
		Afterwards, we added small
		Gaussian white noise\footnote{Without the noise, the
		multi-information rate would be infinite.}. The model was designed to generate
		samples which share many properties with natural images. In particular,
		the samples are \textit{approximately} scale-invariant and share very similar
		marginal and second-order statistics. Many of the difficult-to-capture
		higher-order correlations found in natural images are believed to be
		caused by occlusions in the image. This dataset should therefore pose
		similar challenges as the set of natural images. We extracted training
		data at three different scales and used the same training procedure as
		above. Clearly, our model has not learned what a circle is. However, it is able to
		reproduce the blotchiness of the original samples.
		This is especially surprising given the small size of the neighborhoods and
		the fact that the basic building block of our model is the Gaussian distribution. Also note
		that our model has no explicit knowledge of occlusions. As expected, destroying the
		higher-order correlations of the samples again leaves us with pink noise-like images (Figure
		\ref{fig:leaves_samples}).

		\subsection{Scale invariance and multi-information rates}
			The multiscale representation lends itself to an investigation of
			the scale invariance property of natural images.
			The statistics of a scale-invariant process are
			invariant under block-averaging and appropriate rescaling to
			compensate for the loss in variance \cite{Lee:2001p7371}. Using the
			notation as above, this would mean that $X^0$ is distributed as $a
			X^1$ for some $a$. This in turn implies that the
			multi-information rate (MIR) should stay constant as a function of
			the scale. Because the MIR is invariant under rescaling with a
			constant factor, we can ignore the rescaling factor $a$.

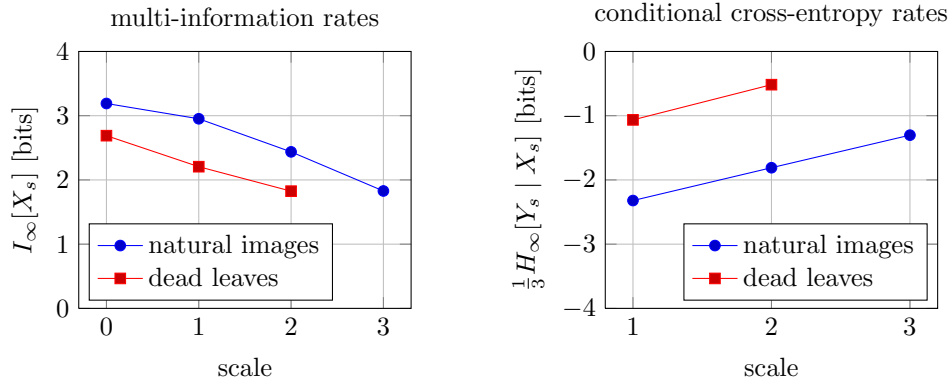
\begin{figure}[t]
	\centering
	\begin{tikzpicture}
		\begin{axis}[
				title={multi-information rates},
				width=6cm,
				height=5cm,
				xlabel={scale},
				ylabel={$I_\infty[X_s]$ [bits]},
				ylabel near ticks,
				ymax=4.0,
				ymin=0.0,
				xmajorgrids=true,
				ymajorgrids=true,
				yminorgrids=true,
				xtick={0,1,2,3},
				ytick={0,1.0,2.0,3.0,4.0},
				yminorticks=true,
				legend style={
					at={(0.03, 0.03)},
					anchor=south west,
					cells={anchor=west}
				},
			]
			\addplot coordinates {
				(0, 3.190)
				(1, 2.952)
				(2, 2.437)
				(3, 1.828)
			};
			\addlegendentry{natural images};

			\addplot coordinates {
				(0, 2.689)
				(1, 2.206)
				(2, 1.825)
			};
			\addlegendentry{dead leaves};
		\end{axis}
		\begin{axis}[
				title={conditional cross-entropy rates},
				at={(7cm,0cm)},
				width=6cm,
				height=5cm,
				xlabel={scale},
				ylabel={$\frac{1}{3} H_\infty[Y_s \mid X_s]$ [bits]},
				ylabel near ticks,
				ymax=0,
				ymin=-4,
				xmajorgrids=true,
				ymajorgrids=true,
				yminorgrids=true,
				xtick={1,2,3},
				ytick={-4,-3,-2,-1,-0},
				yminorticks=true,
				legend style={
					at={(0.97, 0.03)},
					anchor=south east,
					cells={anchor=west}
				},
			]
			\addplot coordinates {
				(1, -2.075 - 0.246)
				(2, -1.564 - 0.246)
				(3, -1.057 - 0.246)
			};
			\addlegendentry{natural images};

			\addplot coordinates {
				(1, -0.819 - 0.246)
				(2, -0.271 - 0.246)
			};
			\addlegendentry{dead leaves};
		\end{axis}
	\end{tikzpicture}
	\caption{
		\textit{Left:} The estimated multi-information rate decreases steadily
		as the scale increases (the resolution decreases). \textit{Right:} The
		conditional cross-entropy rate increases with scale. This shows that
		the van Hateren dataset \cite{vanHateren:1998p7367}
		is generally not scale-invariant. A very similar behavior is shown by images
		created with an occlusion based model \cite{Lee:1999p7586}.
	}
	\label{fig:entropy_rates}
\end{figure}

			We estimated the multi-information rate of the van Hateren
			dataset with the cross-MIR of our model (Figure
			\ref{fig:entropy_rates}). The steady decrease of the information
			rate indicates that the statistics of images taken from the van
			Hateren dataset are not fully scale-invariant. A consequence of a
			smaller MIR is that pixels are more difficult to predict from
			neighboring pixels.

			The difference in cross-MIR could also be caused by the fact that we
			are using a slightly different model at the largest scale than for modeling the image details at
			the lower scales. This problem is not shared by the conditional
			entropy rates plotted on the right of Figure
			\ref{fig:entropy_rates}. If scale invariance is given, the
			distribution over the high-resolution information $Y_s$ and the
			low-resolution information $X_s$ should not change with scale $s$,
			subject to proper rescaling. Since we are using the same model to
			model the relationship between $X_s$ and $Y_s$ for all $s$, the
			estimated entropy rates should stay constant even if our model
			performed poorly. Our results are consistent with the findings of Wu
			et al., who showed that many natural images are more difficult to
			compress at larger scales and argued that the entropy rate of
			natural images has to increase with scale \cite{Wu:2008p7617}.

			Using an estimate of the marginal entropy of $1.57$ bits, we
			arrive at an estimated multi-information rate of $3.44$ bits per
			pixel for the van Hateren dataset (Figure \ref{information_rates}).
			This is approximately $0.18$ bits better than the current
			best estimate for natural images~\cite{Hosseini:2010p7310} and
			$0.04$ bits better than our result obtained without using the
			multiscale representation. 

			Since the true MIR of natural images is unknown, this increase in performance does not
			tell us how much closer we got to capturing all correlations of natural images. It also
			does not reveal in which way the model has improved compared to other models. Samples
			and statistical tests can give us an indication. Figure \ref{fig:comparison} shows
			samples drawn from models suggested by Domke et al. \cite{Domke:2008p7554} and Hosseini
			et al. \cite{Hosseini:2010p7310}, as well samples drawn using the extensions presented in
			this paper. The substantial change in the appearance of the samples suggests that even
			the increase from 3.40 bits to 3.44 bits is significant.

			The joint statistics of the responses of two filters applied at
			different locations in an image are known to change in certain ways
			as a function of their spatial separation and are difficult to
			reproduce \cite{Sinz:2009p7264}. We apply a vertically oriented
			Gaussian derivative filter at two vertically offset locations and
			record their responses. After whitening, the filter responses are
			approximately $L_p$-spherically symmetric. We therefore fit an
			$L_p$-spherically symmetric model with a radial Gamma distribution
			to the responses and, at every distance, record the parameter $p$ of
			the model's norm. Since the marginal distribution of each filter
			response is highly kurtotic and the responses become more
			independent as the filter distance increases, the joint histogram
			becomes more and more star shaped. This is expressed in the optimal
			value for $p$ becoming smaller and smaller. As plotted in Figure
			\ref{information_rates}, the behavior of the optimal $p$ is not
			well reproduced using a single scale but is captured by our
			multiscale model.

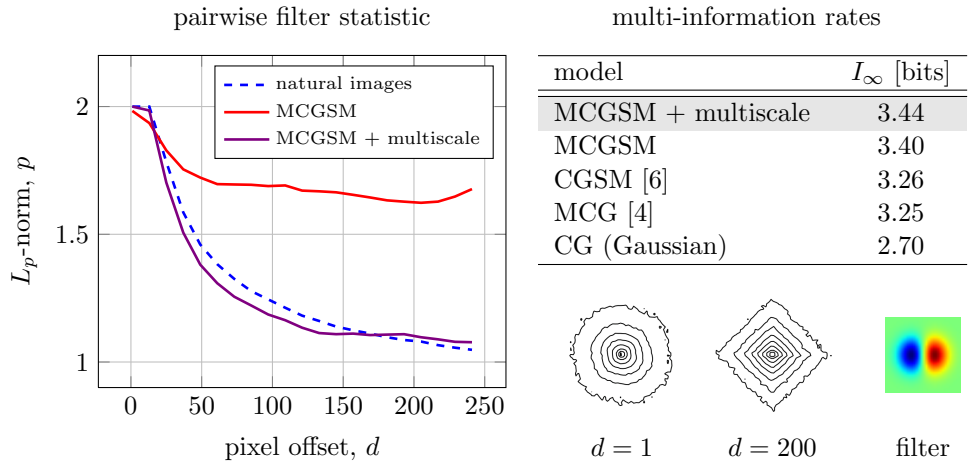
\begin{figure}[t]
	\centering
	\newcommand\T{\rule{0pt}{2.1ex}}
	\newcommand\B{\rule[-0.8ex]{0pt}{0pt}}
	\begin{minipage}{7cm}
		\begin{tikzpicture}
			\begin{axis}[
					width=7cm,
					height=6cm,
					yminorticks=true,
					xmajorgrids=true,
					ymajorgrids=true,
					xlabel={pixel offset, $d$},
					ylabel={$L_p$-norm, $p$},
					ylabel near ticks,
					title={pairwise filter statistic},
					ymin=0.9,
					ymax=2.2,
					legend style={
						legend pos=north east,
						font={\scriptsize},
						cells={anchor=west},
					},
				]

				\addplot [mark=none, blue, dashed, line width=1] coordinates {
					(1, 1.99983558324) +- (0, 3.5555592649e-17)
					(13, 1.99983558324) +- (0, 3.5555592649e-17)
					(25, 1.78333859539) +- (0, 0.00312176144792)
					(37, 1.58567907486) +- (0, 0.00269583682271)
					(49, 1.45975806108) +- (0, 0.00238290247523)
					(61, 1.3828177557) +- (0, 0.00203722064838)
					(73, 1.32549434781) +- (0, 0.00163902944718)
					(85, 1.27755059757) +- (0, 0.00194955668463)
					(97, 1.2450187164) +- (0, 0.00142155313972)
					(109, 1.21253958347) +- (0, 0.00170920949747)
					(121, 1.18149451584) +- (0, 0.00153122146583)
					(133, 1.16067356148) +- (0, 0.00155084722804)
					(145, 1.13874286087) +- (0, 0.00131233994642)
					(157, 1.12408787517) +- (0, 0.00130157390558)
					(169, 1.1114033581) +- (0, 0.00143247394478)
					(181, 1.09704728767) +- (0, 0.00175317459946)
					(193, 1.08605824518) +- (0, 0.0014149249156)
					(205, 1.08009733935) +- (0, 0.00167703619715)
					(217, 1.06599011787) +- (0, 0.00136687315063)
					(229, 1.05603803351) +- (0, 0.00121926426652)
					(241, 1.04773478069) +- (0, 0.0011908053572)
				};
				\addlegendentry{natural images};

				\addplot [mark=none, red, line width=1] coordinates {
					(1, 1.98290641606) +- (0, 0.0027116217611)
					(13, 1.9353101648) +- (0, 0.00300510289149)
					(25, 1.82840627731) +- (0, 0.00289000093962)
					(37, 1.75413771871) +- (0, 0.00275017201856)
					(49, 1.72197754091) +- (0, 0.00209802815802)
					(61, 1.69636907803) +- (0, 0.00221918273496)
					(73, 1.69495145439) +- (0, 0.00202000835407)
					(85, 1.69378656105) +- (0, 0.00265139663175)
					(97, 1.68868997576) +- (0, 0.0027540120561)
					(109, 1.69116099921) +- (0, 0.00200738623753)
					(121, 1.67094415553) +- (0, 0.00175480341979)
					(133, 1.66821726169) +- (0, 0.00196112420859)
					(145, 1.66427809016) +- (0, 0.00216345153449)
					(157, 1.65447352276) +- (0, 0.00230257785123)
					(169, 1.64410235066) +- (0, 0.00225284416796)
					(181, 1.63280109131) +- (0, 0.00205324429738)
					(193, 1.62772796912) +- (0, 0.00250768517435)
					(205, 1.6232825957) +- (0, 0.00201536888025)
					(217, 1.62786846464) +- (0, 0.00206701916532)
					(229, 1.64767444143) +- (0, 0.00258320810255)
					(241, 1.67695198759) +- (0, 0.00217817239347)
				};
				\addlegendentry{MCGSM};

				\addplot [mark=none, violet, line width=1] coordinates {
					(1, 1.99983558324) +- (0, 3.5555592649e-17)
					(13, 1.98454120471) +- (0, 0.00274694735864)
					(25, 1.70247436074) +- (0, 0.00268604881569)
					(37, 1.50635069567) +- (0, 0.00222272967477)
					(49, 1.38109013902) +- (0, 0.00164656178913)
					(61, 1.30842080378) +- (0, 0.00157488362539)
					(73, 1.25566025988) +- (0, 0.00163334347641)
					(85, 1.22119883928) +- (0, 0.00171018214288)
					(97, 1.18598459167) +- (0, 0.00127184042563)
					(109, 1.16372717316) +- (0, 0.00138160321902)
					(121, 1.13424262353) +- (0, 0.00145481544215)
					(133, 1.11335082173) +- (0, 0.0011125706254)
					(145, 1.10897851959) +- (0, 0.000795696106151)
					(157, 1.1111607893) +- (0, 0.00130175877303)
					(169, 1.10544633361) +- (0, 0.0012101327215)
					(181, 1.10727168408) +- (0, 0.0011139614948)
					(193, 1.10889556994) +- (0, 0.00131154990865)
					(205, 1.09694281558) +- (0, 0.00117140560466)
					(217, 1.08877811759) +- (0, 0.00140420936112)
					(229, 1.07903497742) +- (0, 0.00128831281244)
					(241, 1.07742365188) +- (0, 0.00100445081216)
				};
				\addlegendentry{MCGSM + multiscale};
			\end{axis}
		\end{tikzpicture}
	\end{minipage}
	\begin{minipage}{6cm}
		\vspace{0.06cm}
		\begin{center}
			\hspace{-0.45cm}multi-information rates
		\end{center}
		\vspace{0.02cm}
		\begin{tabular}{lc}
			\hline
			model \T\B & $I_\infty$ [bits] \\
			\hline
			\hline
			\rowcolor[gray]{.9}
			MCGSM + multiscale \T\B & 3.44 \\
			MCGSM \T\B & 3.40 \\
			CGSM \cite{Hosseini:2010p7310} \T\B & 3.26 \\
			MCG \cite{Domke:2008p7554} \T\B & 3.25 \\
			CG (Gaussian) \T\B &  2.70 \\
			\hline
		\end{tabular}
		\vspace{0.04cm}
		\hspace{-0.45cm}
		\begin{tikzpicture}
			\pgfdeclareimage[width=2cm,height=2cm]{dist001}{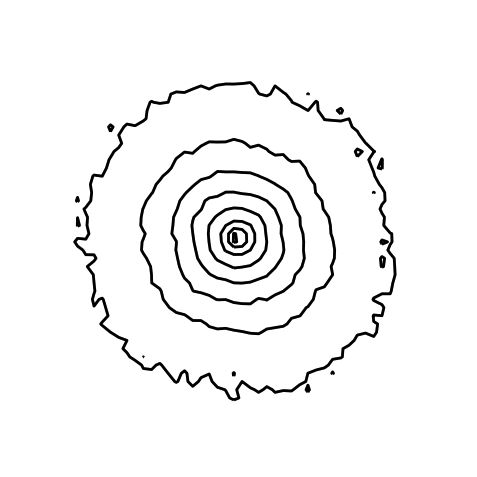}
			\pgfdeclareimage[width=2cm,height=2cm]{dist200}{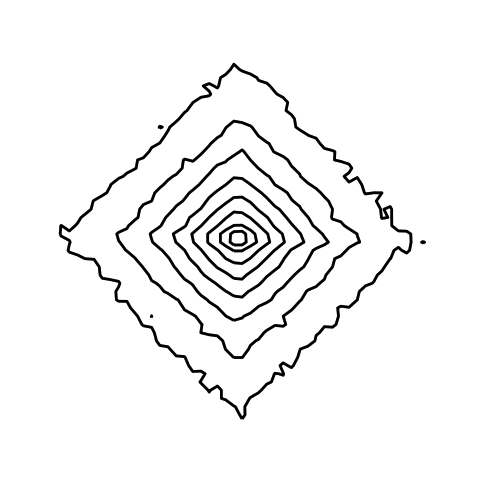}
			\pgfdeclareimage[width=1cm,height=1cm]{gdfilter}{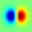}
			\node at (5cm, 3.5cm) {\pgfuseimage{dist001}};
			\node at (5cm, 2.28cm) {$d = 1$};
			\node at (7cm, 3.5cm) {\pgfuseimage{dist200}};
			\node at (7cm, 2.28cm) {$d = 200$};
			\node at (9cm, 3.5cm) {\pgfuseimage{gdfilter}};
			\node at (9cm, 2.28cm) {filter};
		\end{tikzpicture}
		\vspace{0.01cm}
	\end{minipage}
	\caption{\textit{Top right:} Multi-information rate estimates of natural
		images obtained using different models. The multiscale representation allows
		us to obtain a somewhat better estimate.
		\textit{Bottom right:} The joint histogram of pairs of Gaussian
		derivative filter responses changes as their spatial
		separation increases. \textit{Left:} $L_p$-spherically symmetric
		distributions were fitted to the filter responses for natural and
		synthetic data. The vertical axis shows a
		maximum likelihood estimate of the parameter $p$.
		The horizontal axis shows the vertical offset between the position of the two
		filters. The plot shows that the multiscale representation
		enables our model to match the statistics of pairwise filter responses over much
		longer distances, which could be one possible explanation for the better
		performance.}
	\label{information_rates}
\end{figure}
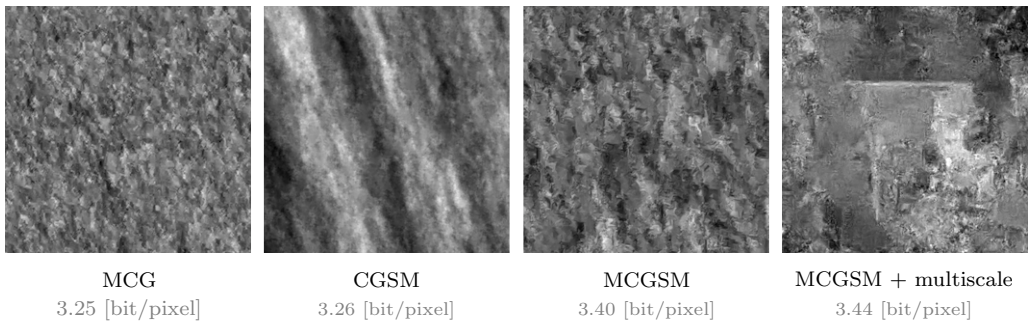
\begin{figure}[t]
	\hspace{-0.2cm}
	\begin{tikzpicture}
		\node at (0, 0) {\includegraphics[width=\textwidth]{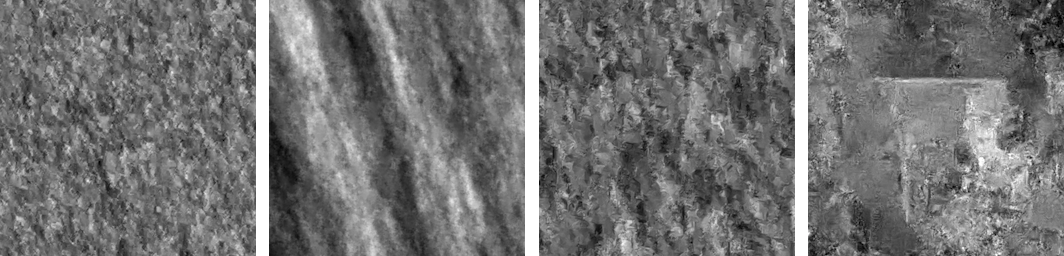}};
		\node at (-5.15, -2) {\footnotesize MCG};
		\node at (-5.15, -2.4) {\footnotesize \color{gray} 3.25 [bit/pixel]};
		\node at (-1.73, -2) {\footnotesize CGSM};
		\node at (-1.73, -2.4) {\scriptsize \color{gray} 3.26 [bit/pixel]};
		\node at (1.73, -2) {\footnotesize MCGSM};
		\node at (1.73, -2.4) {\scriptsize \color{gray} 3.40 [bit/pixel]};
		\node at (5.15, -2) {\footnotesize MCGSM + multiscale};
		\node at (5.15, -2.4) {\scriptsize \color{gray} 3.44 [bit/pixel]};
	\end{tikzpicture}
	\caption{
		\textit{From left to right:} Samples from a mixture of conditional Gaussians \cite{Domke:2008p7554}
		(5x5 neighborhoods, 5 components including means), a conditional Gaussian scale mixture
		\cite{Hosseini:2010p7310} (7x7 neighborhoods, 7 scales), a mixture of conditional Gaussian scale mixtures and the multiscale model. The appearance
		of the samples changes drastically from model to model, indicating that the seemingly small improvements
		in bit/pixel are in fact substantial.
	}
	\label{fig:comparison}
\end{figure}

	\section{Conclusion}
		We have shown how to use directed models in combination with
		multiscale representations in a way which allows us to still evaluate
		the model in a principled manner. To our knowledge, this is the only
		multiscale model for which the likelihood can be evaluated. Despite the model's computational
		tractability, it is able to learn interesting higher-order correlations
		from natural images and yields state-of-the-art performance when
		evaluated in terms of the multi-information rate. In contrast to the
		directed model applied to images at a single scale, the model also
		reproduces the pairwise statistics of filter responses over long distances.
		Here, we only used a simple multiscale representation. Using more sophisticated
		representations might lead to even better models.

	\bibliographystyle{unsrt}
	\small
	\bibliography{references}
\end{document}